%% file: evw2020paper.tex
\documentclass[runningheads]{llncs}
\usepackage{graphicx}
\usepackage{tikz}
\usepackage{comment}
\usepackage{amsmath,amssymb} % define this before the line numbering.
\usepackage{color}

\usepackage{textcomp}
\input{math_commands.tex}

\usepackage{algorithm}
\usepackage{algorithmic}
\usepackage{wrapfig}
\usepackage{wrapfig,lipsum,booktabs}

\begin{document}
% \renewcommand\thelinenumber{\color[rgb]{0.2,0.5,0.8}\normalfont\sffamily\scriptsize\arabic{linenumber}\color[rgb]{0,0,0}}
% \renewcommand\makeLineNumber {\hss\thelinenumber\ \hspace{6mm} \rlap{\hskip\textwidth\ \hspace{6.5mm}\thelinenumber}}
% \linenumbers
\pagestyle{headings}
\mainmatter
\def\ECCVSubNumber{24}  % Insert your submission number here

\title{Subtensor Quantization for Mobilenets} % Replace with your title

% INITIAL SUBMISSION 
\begin{comment}
\titlerunning{EVW-20 submission ID \ECCVSubNumber} 
\authorrunning{EVW-20 submission ID \ECCVSubNumber} 
\author{Anonymous EVW submission}
\institute{Paper ID \ECCVSubNumber}
\end{comment}
%******************

% CAMERA READY SUBMISSION
%\begin{comment}
\titlerunning{Subtensor Quant for Mobilenets}
% If the paper title is too long for the running head, you can set
% an abbreviated paper title here
%
\author{Thu Dinh \and
Andrey Melnikov \and
Vasilios Daskalopoulos \and
Sek Chai}
\authorrunning{T. Dinh et al.}
% First names are abbreviated in the running head.
% If there are more than two authors, 'et al.' is used.
%
\institute{Latent AI, Princeton NJ 08558, USA}
%\email{first@latentai.com}
%\url{http://www.latentai.com}
%\end{comment}
%******************
\maketitle

\begin{abstract}
Quantization for deep neural networks (DNN) have enabled developers to deploy models with less memory and more efficient low-power inference. However, not all DNN designs are friendly to quantization. For example, the popular Mobilenet architecture has been tuned to reduce parameter size and computational latency with separable depthwise convolutions, but not all quantization algorithms work well and the accuracy can suffer against its float point versions. In this paper, we analyzed several root causes of quantization loss and proposed alternatives that do not rely on per-channel or training-aware approaches. We evaluate the image classification task on ImageNet dataset, and our post-training quantized 8-bit inference top-1 accuracy in within 0.7\% of the floating point version.
\end{abstract}

\section{Introduction}
%\vspace{-8pt}
Quantization is a crucial optimization for DNN inference, especially for embedded platforms with very limited budget for power and memory. Edge devices on mobile and IoT platforms often rely fixed-point hardware rather than more power-intensive floating point processors in a GPU. Through quantization, parameters for DNN can be converted from 32-bit floating point (FP32) towards 16-bit or 8-bit models, with minimal loss of accuracy. While significant progress has been made recently on quantization algorithms, there is still active research to understand the relationships among bit-precision, data representation, and neural model architecture.

One of the main challenges for quantization is the degrees of non-linearity and inter-dependencies among model composition, dataset complexity and bit-precision targets. Quantization algorithms assign bits for the DNN parameters, but the best distribution depends on range of values required by each model layer to represent the learnt representation. As the model grows deeper, the complexity of analysis increases exponentially as well, making it challenging to reach optimal quantization levels. 

In this short paper, we present a case-study of a state-of-the-art mobilenetv2 \cite{Sandler2018} architecture from a quantization perspective. We propose a new algorithmic approach to enhance asymmetric quantization to support depthwise convolutions. In comparison, current research has steered towards more complicated quantization schemes that require more memory and hardware support. Using our approach instead, the overall DNN inference uses a simpler quantization scheme and still maintain algorithm accuracy; all achieved without retraining the model. To the best of our knowledge, we offer the following contributions in this paper: (1) new algorithmic approach for enhanced asymmetric quantization, and (2) comparison of accuracy, memory savings, and computational latency.

%Our goal is to change the perception that DNN quantization is a solved problem. Specifically, for post-training quantization, we show that current approaches to handle depthwise convolution layers could result in solutions that not optimal from a memory and latency perspective. Through analysis of quantization effects using range and distributions of the DNN parameter values, we were able to develop a new algorithm that offers further DNN compression without reliance on hardware (memory and compute) support. Our work is enabled by a combined workflow integrates a compiler to generate native binary on embedded platforms [LEIP].
%
%This paper is organized as follows: in Section 2, we present a summary of recent related research in DNN quantization. In Section 3, we briefly define our quantization approaches with highlights on the proposed algorithm. In Section 4, we describe our simulation setup and early results with associated analysis that evaluates our approach. Finally, in Section 5, we present our conclusions and discuss our future work in this space.
%
\section{Related Work}
%\vspace{-8pt}
Most post-training quantization (PTQ) algorithms are based on finding the mapping between floating point and an evenly spaced grid that can be represented in integer values. The goal is to find the scaling factor (e.g. max and min ranges) and zero-point offset to allow such mapping \cite{Guo2018,krishna2018,nayak2019}. Similar to \cite{Nagel2019}, our approach is a data-free quantization, whereby no training data or backprogation is required. Additional methods have been proposed to deal specifically with various elements of a DNN layer such as  the bias tensors \cite{Nagel2019,Finkelstein2019}. The goal remains the same for all algorithms: reduce bit-precision and memory footprint while maintaining accuracy.

There are other methods for DNN compression including, network architecture search and quantization aware training \cite{Parajuli2018}. Some training aware methods may binarize or ternarize, result in in DNNs that operate with logical operations rather than multiplications and additions. These approaches are in a different algorithmic category and are not considered for discussion in this short paper.

\section{Proposed Algorithm}
%\vspace{-10pt}

\begin{wrapfigure} {r}{0.45\textwidth}
  \centering
    %\vspace{-25pt}
    \includegraphics[width=0.45\textwidth]{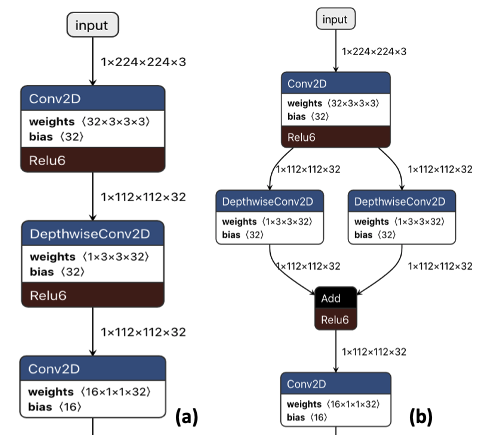}
    %\vspace{-20pt}
  \caption{Example DNN model snippet. (a) Original, (b) Proposed subtensor optimization for quantization.}
  %\vspace{-15pt}
  \label{fig:graph}
\end{wrapfigure}

In this paper, we use a per-tensor asymmetric PTQ scheme where the min/max in the float range is mapped to the min/max of the integer range \cite{nayak2019}. This is performed by using a zero-point (also called quantization bias, or offset) in addition to the scale factor. For per-channel approach, the quantization scheme is done separately for each channel in the DNN \cite{krishna2018}. Convolutional and dense layers consist of a significant number of channels. Instead of quantizing all of them as tensors, per-channel quantization can be used to quantize each channel separately to provide accuracy improvements.

We propose a new enhancement for quantizing tensors by splitting the tensors into subtensors, in order to achieve both high resolution and range in integer representation. Fig.\ref{fig:graph} shows a diagram of the original DNN graph and our proposed subtensor modification. In our approach, subtensors are better able to be quantized when done separately (e.g. splitting a problem into smaller parts makes it easier and more feasible to find the optimal mapping using asymmetric quantization). We then apply bias correction that includes weight clipping analysis and cross-layer equalization approach.

In doing so, we are increasing the bit-precision capacity of the DNN without changing the overall architecture. Previous per-tensor approaches are not able to achieve high algorithmic accuracy because there is simply not enough bits to encode both the resolution and range of a single tensor. Specifically, per-channel approaches allocates more bits along DNN channels, which requires memory access patterns that are different and less practical for most CPUs hardware. 

Tensor splitting is analogous to channel splitting in \cite{Zhao2019}, except that their approach duplicate tensors rather than splitting and optimizing subtensors separately. They re-scale of each duplicate tensors in order to achieve reduce floating point ranges, which loses resolution on critical portions floating point range and consequently, overall accuracy suffers. 

\section{Results and Analysis}
%\vspace{-8pt}
For evaluation, we used the workflow \cite{LEIP} that builds upon a TensorFlow framework to implement all of the quantization approaches. For the DNN architecture, we use a baseline mobilenetv2 model, trained with imagenet2012 dataset (1000 classes). The pre-trained model is then quantized and then compiled to generate either a native binary that can run independently or with a TFLite runtime. This workflow allows quick evaluation on embedded because the compiler targets optimal code for target embedded hardware. 

\begin{wraptable}{r}{6.5cm}
\small
%\vspace{-20pt}
\centering
\caption{PTQ Results}\label{table:PTQ}
\begin{tabular}{l|l|l|l}
		\hline
		\textbf{MobileNetv2} & Accuracy & Latency & Size \\
		  & (\%) & (ms) & (kB) \\  \hline
		Baseline FP32 \cite{mnetv2} & 71.8 & 20.4 & 13651 \\ \hline
	    Per-Channel \cite{krishna2018} & 69.8 & 11.5 & 4056 \\ \hline
	    Training Aware \cite{mnetv2} & 70.7 & 8.2 & 3494 \\ \hline
	    Bias Fine Tune \cite{Finkelstein2019}& 70.6 & 8.2 & 3494 \\ \hline
	    Proposed \cite{LEIP} & 71.12 & 9.1 & 3508  \\ \hline
\end{tabular}
%\vspace{-10pt}
\end{wraptable} 

%{describe the table comparisons}
Table \ref{table:PTQ} show quantization comparison results for mobilenetv2. There are many variants of pre-trained models, and we use a baseline floating-point (FP32) model and apply various PTQ algorithms. The training-aware results are provided only as reference. We show that our proposed algorithm can reach within 0.7\% of the baseline. Using the proposed subtensor quantization, we are trading tensor processing time to achieve higher algorithm accuracy. We have found that the additional processing time for the subtensors for the additional layer (the ADD layer) is minimal compared to the overall DNN processing time (only 4\% of baseline). The additional overhead in memory size is also negligible (only 0.1\% of baseline). Our PTQ approach using subtensors achieve a high level of accuracy, while maintaining flexibility, file size and processing speed. 

In comparison, using per-channel quantization, each tensor is quantized separately for each channel \cite{krishna2018}. A channel with many outlier weights will be quantized separately, resulting in improved accuracy. However, this method requires a set of quantization parameters (min-val, max-val, zero-point, scale) for each channel. Computationally, the model has to perform quantization, multiply, add, dequantization separately for each channel, for each tensor. This results in a much slower inference rate due to the more complicated memory access. Moreover, this technique of quantization is not accepted by some compilers or hardware. Specifically, our proposed subtensor approach is more hardware agnostic as the tensor memory access patterns are similar throughout the DNN.

The mobilenet architecture, with seperable depthwise convolutions can exhibit a significant shift in the mean activation value after quantization. This shift is a result of quantization errors being unbalanced, and as such, bias correction improved overall quantized model performance. However, bias correction does not achieve near FP32 performance on its own. As shown in Table \ref{table:PTQ}, subtensor level optimization improves the overall performance and brings algorithmic accuracy back to the floating point baseline.

\section{Conclusion}
%\vspace{-8pt}
Our goal is to change the perception that DNN quantization is a solved problem. Specifically, for post-training quantization, we show that current approaches to handle depthwise convolution layers could result in solutions that are not optimal. Through analysis of quantization effects using range and distributions of the DNN parameter values, we propose a new algorithm based on subtensor optimization that offers further DNN compression without reliance on hardware (memory and computation) support. Our work is enabled by a combined workflow that integrates a compiler to generate native binary on embedded platforms \cite{LEIP}. We show performance within 0.7\% of accuracy for a mobilenetv2 architecture, quantized from a floating point (FP32) baseline to integer 8-bits precision. Future work includes enhancements in the selection of subtensors to further reduce quantization errors.

% ---- Bibliography ----
%
% BibTeX users should specify bibliography style 'splncs04'.
% References will then be sorted and formatted in the correct style.
%
\bibliographystyle{splncs04}
%\vspace{-6pt}
\bibliography{egbib}
\end{document}

%% file: math_commands.tex
\usepackage{amsmath,amsfonts,bm}

\def\eqref#1{equation~\ref{#1}}
\def\1{\bm{1}}

\newcommand{\test}{\mathcal{D_{\mathrm{test}}}}

\DeclareMathAlphabet{\mathsfit}{\encodingdefault}{\sfdefault}{m}{sl}
\SetMathAlphabet{\mathsfit}{bold}{\encodingdefault}{\sfdefault}{bx}{n}